%% file: Main-sn.tex
\documentclass[pdflatex,sn-apa]{sn-jnl}

\usepackage{graphicx}%
\usepackage{multirow}%
\usepackage{amsmath,amssymb,amsfonts}%
\usepackage{amsthm}%
\usepackage{mathrsfs}%
\usepackage[title]{appendix}%
\usepackage{booktabs}%

\usepackage{pifont}
\usepackage{adjustbox}
\usepackage{listings}
\usepackage{makecell}
\usepackage{tcolorbox} 

\newcommand{\xmark}{\ding{55}}

\theoremstyle{thmstyleone}%
\theoremstyle{thmstyletwo}%

\theoremstyle{thmstylethree}%

\begin{document}

\title[Human-like Affective Cognition in Foundation Models]{Human-like Affective Cognition in Foundation Models}

\vspace{-5mm}


\author*[1]{\fnm{Kanishk} \sur{Gandhi}}\email{kanishk.gandhi@stanford.edu}

\author[1]{\fnm{Zoe} \sur{Lynch}}\email{zlynch@stanford.edu}

\author[2]{\fnm{Jan-Philipp} \sur{Fr\"anken}}\email{janphilipp.franken@gmail.com}

\author[1]{\fnm{Kayla} \sur{Patterson}}\email{kpatterson@alumni.stanford.edu
}

\author[1]{\fnm{Sharon} \sur{Wambu}}\email{swambu@stanford.edu
}

\author[2]{\fnm{Tobias} \sur{Gerstenberg}}\email{gerstenberg@stanford.edu
}

\author[3]{\fnm{Desmond C.} \sur{Ong}}\email{desmond.c.ong@gmail.com
}

\author[1,2]{\fnm{Noah D.} \sur{Goodman}}\email{ngoodman@stanford.edu}

\affil*[1]{\orgdiv{Computer Science}, \orgname{Stanford University}, \orgaddress{\street{450 Jane Stanford Way},\\ \city{Stanford}, \postcode{94305}, \state{CA}, \country{USA}}}
\affil[2]{\orgdiv{Psychology}, \orgname{Stanford University}, \orgaddress{\street{450 Jane Stanford Way},\\ \city{Stanford}, \postcode{94305}, \state{CA}, \country{USA}}}
\affil[3]{\orgdiv{Psychology}, \orgname{The University of Texas}, \orgaddress{\street{110 Inner Campus Drive},\\ \city{Austin}, \postcode{78712}, \state{TX}, \country{USA}}}

\vspace{-10mm}
\abstract{
\input{sections/00_abstract}
}
\vspace{-5mm}

\keywords{Affective Cognition, Large Language Models, LLMs, Affective Computing, Cognitive Science}
\vspace{-5mm}
\maketitle

\label{sec:Intro}
\section{Introduction}\label{sec:Results}

\input{sections/01_intro}

\section{Results}\label{sec:Results}
\input{sections/02_results}

\section{Discussion}\label{sec:Discussion}
\input{sections/03_discussion}


\bmhead{Acknowledgements}
We would like to thank Michael Bernstein, and the members of the Computation and Cognition Lab for their support and feedback.
This work was supported by the Stanford Human-Centered Artifical Intelligence (HAI) - Google grant.
This material is based upon work supported by the National Science Foundation under Award No. 2443038 to D.C.O., and an Expeditions Grant, Award Number (FAIN) 1918771 to N.D.G.
Any opinions, findings and conclusions or recommendations expressed in this material are those of the author(s) and do not necessarily reflect the views of the National Science Foundation.
\backmatter

\newpage
\section{Materials and Methods}

The complete methods and materials for reproduction are available at: 
\\\href{https://github.com/kanishkg/affective-cog}{https://github.com/kanishkg/affective-cog}

\noindent
Here we detail the main prompts and parameters used for evaluation.

\subsection*{Prompts and Parameters}
For inference, we use a temperature of 0.0 and a top-p value of 0.9. For GPT-4, we used \texttt{gpt-4-1106}. For Claude-3.5, we used \texttt{claude-3.5-sonnet-20240620}. For Gemini, we used \texttt{gemini-1.5-pro-002}. Please see \autoref{exp:0shot} for the 0-shot prompt, and \autoref{exp:0shotcot} for the 0-shot chain-of-thought prompt that was used for the text-only stimuli. The prompts for the stimuli with expression are presented in \autoref{exp:0shotx}, for 0-shot and \autoref{exp:0shotcotx} for 0-shot chain-of-thought.
\label{asec:prompt}

\subsection*{Distributional distance between responses}
\input{sections/14_jsd}

\input{sections/sup_figs}

\newpage

\begin{appendices}




\end{appendices}


\bibliography{ref}

\end{document}

%% file: sections/00_abstract.tex

Understanding emotions is fundamental to human interaction and experience, and also for AI models that interact with people. Such understanding goes beyond recognizing emotions from facial expressions or language; it involves reasoning over how people subjectively make sense of their experienced situations given their beliefs and goals, and integrating multiple sources of information. Starting from psychological theory on affective cognition, we introduce a principled evaluation framework for assessing such reasoning in AI models. Using this framework, we generate 1,280 diverse scenarios systematically varying the relationships between appraisals, emotions, expressions, and situation outcomes, and evaluated the responses of three Large Language Models (GPT-4, Claude-3.5, Gemini-1.5) and humans (N = 567). Our results show that these models tend to agree with human intuitions, matching and in some cases exceeding interparticipant agreement, and that all models' performances increase using chain-of-thought reasoning. This suggests that Large Language Models seem to have acquired a conceptual understanding of, and are able to reason about, human emotions.

%% file: sections/01_intro.tex
Having emotions is fundamental to being human, and our emotions are shaped by how we view our experiences. 
For instance, consider Amy, a high school student who is applying to college. She wants to attend a local state college, but her parents want her to go to a private liberal arts college. 
If she is admitted to the private liberal arts college and not the local state college, she is disappointed; but she would be happy if she gets into the local state college. A friend who observes Amy's disappointment and knows her preferences can infer her rejection from the state college. Similarly another friend who doesn't know her preference but observes the outcome and her emotion may infer which colleges she wanted to go to.
This ability to understand others' emotions in the context of their mental states, known as affective cognition ~\citep{ong2015affective, saxe2017formalizing}, allows people to gain deeper insights into others' thoughts and experiences ~\citep{de2014reading, wu2018rational, houlihan2023emotion}, fostering better connections and interactions.  
This fundamentally human capacity 
is the ability that allows us to be understanding friends, empathetic counselors, and compassionate partners.

Recent advances in foundation models
\citep{Anthropic2024,touvron2023llama,achiam2023gpt,team2023gemini} 
have led to AI becoming an increasing part of our daily interactions \citep{demszky2023using, bommasani2021opportunities, tamkin2021understanding}. 
Indeed, recent statistics published by Anthropic \citep{anthropic2025affective} and OpenAI \citep{phang2025investigating, chatterji2025people} suggests that a small percentage of conversations with Claude (about 3\%) and chatGPT are affective in nature---e.g., users seeking advice on relationships, career development, self-reflection, managing loneliness. While this percentage may seem small, this translates to potentially millions of affective conversations a day for these ``general purpose" models; and the numbers are likely higher for models labeled as ``empathic" or ``emotionally intelligent" (e.g., Inflection's Pi, Hume's EVI), or personas built on top of general purpose models. 
It is thus important to measure how well models understand people, including our emotions. If AI assistants and companions do not understand the nuances of common emotions such as sadness, joy, or frustration, they will be fundamentally limited in their ability to connect with us \citep{picard2000affective}. 
This raises an important question: Can modern AI models understand emotions, making the same inferences that \emph{humans} do? 

Crucially, emotion understanding goes beyond emotion recognition from faces (facial expression recognition; \citealp{kleinsmith2012affective,shan2009facial,li2020deep}), text (sentiment analysis; \citealp{medhat2014sentiment,zhang2018deep, rathje2024gpt}), or other modalities like voice. Conceptually, these emotion recognition tasks are \emph{perception} tasks---they attempt to recover some ``ground truth" label solely from information only in the stimulus. 
By contrast, people employ more complex types of reasoning to understand emotions. 
For instance, affective cognition requires a rich, causal intuitive theory of how emotions are related to mental states and contexts \citep{ong2019computational, doan2025emotion, saxe2017formalizing}. Central to connecting mental states with emotions is an understanding of how people evaluate events based on their prior expectations, beliefs and desires, a process called cognitive appraisal  \citep{ellsworth2003appraisal, scherer2001appraisal, skerry2015neural, weiner1985attributional, yeo2024meta}. For example, when Amy receives news of her rejection from the local state college, her emotions reflect not just the outcome but her interpretation of the outcome. This \emph{appraisal} is critical to how a situation leads to an emotion. Given the same outcome, a different appraisal can lead to a different emotion---if Amy thought she would certainly get into the college on a second try, she would not feel as disappointed, and instead would feel motivated to try again.
In this example, Amy's emotions were a consequence of her appraising the outcome (a rejection) as being incongruent with her goals.
There are many other dimensions of appraisal which could lead to other nuances in her emotions. For example, if Amy thought that she had a great deal of influence over the outcome of the college admission process and that it was mostly her application that caused a rejection, she might feel regret, or frustration.

We do not yet know the extent to which recent AI models, especially foundation models, can reason about human emotions. Some recent work has investigated the ability of LLMs like GPT-4 to infer people's emotions and appraisals from vignettes \citep{broekens2023fine, tak2023gpt, tak2024gpt, yongsatianchot2023investigating, zhan2023evaluating}, and finding some promising initial results. However, these studies lack a principled way of (i) defining a taxonomy of different types of affective inferences, and (ii) systematically benchmarking those inferences.

\begin{figure}[!tbp]
    \centering
\includegraphics[width=1.0\textwidth]{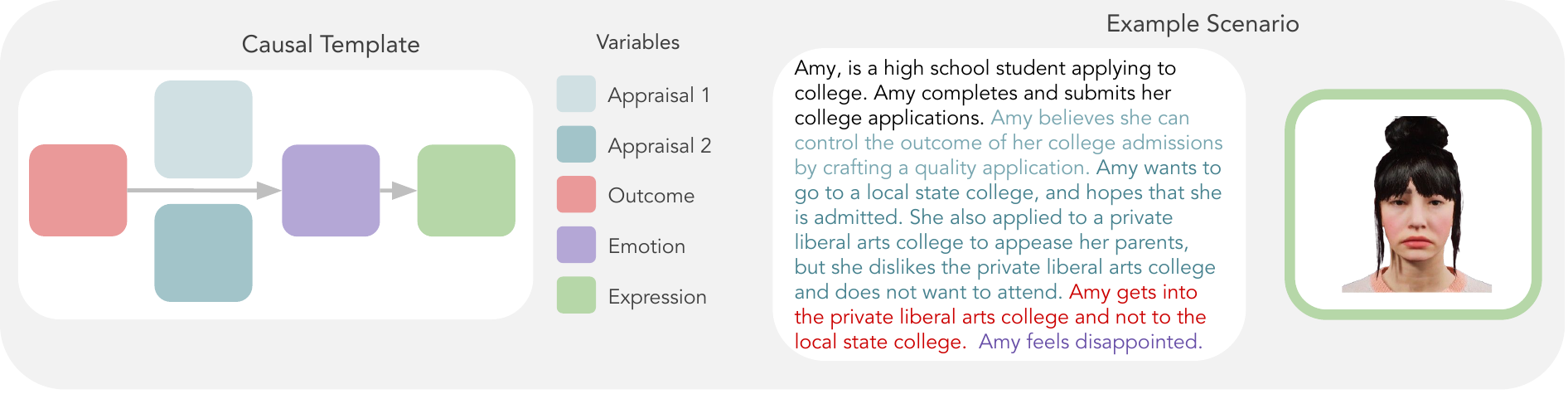}
    \caption{\textbf{Causal Template for generating affective scenarios and an Example Scenario.} (left) The causal template used to generate stimuli for testing affective inferences. Experiments 1a and 1b use the left four text-only causal factors, while Experiments 2a and 2b use all five factors including the Expression factor (represented as an image). (right) An example scenario generated with our causal template for affective inferences. The color of the text indicates the causal variable associated with it.}
    \label{fig:causal}
\end{figure}

To address these gaps, 
we propose a framework to generate structured tests for affective cognition, and compare how recent AI models perform. Our items are built synthetically, extending techniques for evaluating language models using language models \citep{gandhi2024understanding, fränken2024procedural,perez2022discovering,perez2022red}. Our approach starts with a strong theoretical grounding, which allows us to exhaustively define various types of inferences within affective cognition. We then use a systematic approach to isolate each of these inferences, and procedurally generate a range of items to test them. The procedural generation of stimuli has 3 stages: 1) Defining an abstract causal template. 2) Populating the template using language models. 3) Composing the stimuli from the populated causal template.

In Stage 1, we start by specifying an abstract causal graph for affective reasoning (\autoref{fig:causal}), grounded in psychological theory \citep{ong2019computational, saxe2017formalizing}, which describes some of the causal relationships between outcomes, appraisals, and emotions \citep{ellsworth2003appraisal, yeo2024meta}. 
In Experiments 1a and 1b, we use a four-factor causal model (\autoref{fig:causal}, ``Causal Template") where an \textbf{outcome} and two \textbf{appraisals} cause an \textbf{emotion}. In Experiments 2a and 2b, we add a fifth factor to the causal model, such that the emotion causes an \textbf{expression}; we create images depicting a facial expression, and which allows us to test multimodal models. 

For the example scenario in \autoref{fig:causal}, we have a background story (applying to college) and what the person's goals and expectations are (e.g., Amy wants to attend the local state college, and that application and results are well within her control). When the \textbf{outcome} occurs (Amy gets rejected from the local state college), the person would make certain \textbf{appraisals} (e.g., she experienced an outcome incongruent with her goal (\emph{goal-congruency}); and she perceives that she had \emph{control} over the outcome). According to cognitive appraisal theory, these specific appraisals should lead her to feel an \textbf{emotion} (like disappointed). In this particular example, and in Experiments 1a and 2a, we consider two appraisal dimensions (\emph{goal-congruency} and \emph{perceived control}), while in Experiments 1b and 2b, we consider \emph{safety} and \emph{expectedness} (see \autoref{tab:emotions}). These two sets of experiments serve as proofs-of-concept, and this paradigm can be extended to cover a wider range of appraisals and emotions (e.g., \citealp{yeo2024meta} identified 47 distinct appraisal dimensions and how they relate to 63 different emotions and affective states).

\begin{table}[htbp]
  \centering  
\caption{\textbf{Appraisal Dimensions and Assigned Emotion Labels.} Appraisal dimensions and their corresponding emotion labels assignment prior to collecting human judgments, based on theory. \checkmark~ indicates a positive value for the appraisal dimension and \xmark~ indicates a negative value. The left table shows the two appraisals studied in Experiments 1a and 2a, while the right table shows the appraisals in Experiments 1b and 2b.
  }

  \begin{tabular}{ccc||ccc}

    \textbf{Goal} & \textbf{Control} & \textbf{Emotion} & \textbf{Safety} & \textbf{Expectedness} & \textbf{Emotion} \\
    \textbf{Congruency} & & (from theory) & & & (from theory) \\
    \midrule
    \checkmark & \checkmark & Joyful & \checkmark & \checkmark & Relieved \\
    \xmark & \checkmark & Frustrated & \xmark & \checkmark & Resigned \\
    \checkmark & \xmark & Grateful & \checkmark & \xmark & Surprised \\\
    \xmark & \xmark & Disappointed & \xmark & \xmark & Devastated \\
    \bottomrule
  \end{tabular}
    \label{tab:emotions}
\end{table}

In Stage 2, once this causal template is specified, we prompt a language model to generate values for these factors. We first have the language model generate a background story, such as: ``Amy is a high school student applying to college.'' We then use the model to generate text corresponding to two (binary) values for each of the appraisal dimensions.
For \emph{goal-congruence}, the model generates ``Amy wants to go to a local state college and not a private liberal arts college.'' and its complementary value ``Amy wants to go to a private liberal arts college and not a local state college.'' For perceived \emph{control} over the outcome, the model would generate ``Amy thinks that she can influence the decision of the admission process.'' and ``Amy thinks that the admission process is mostly random and she has little influence over the decision.'' We also have the model generate two possible outcomes, ``Amy is accepted at the local state college and rejected from the private liberal arts college'' and ``Amy is accepted at the private liberal arts college and rejected from the local state college.''. Based on the specific appraisal factors and appraisal theory, we define four emotion values that the person in the story might feel (\autoref{tab:emotions}) --- for example, ``Amy feels disappointed''. Finally, to create the image stimuli we use in Experiments 2a and 2b, we match each emotion with facial expressions that are defined using Facial Action Units \citep{ekman1978facial} and rendered in Unity (\autoref{fig:causal}). 

Importantly, the language model \emph{does not} have to make affective inferences while generating values for the variables in the template; it simply needs to follow prompts to populate the causal template.


Finally, in Stage 3, we compose the text completions to produce individual stimuli items. Populating the abstract causal template for a single background story yields eight possible filled-out scenarios (2 values for each of 2 appraisals, and 2 values for the outcome; the emotion is determined from these other variables). The Example Scenario in \autoref{fig:causal} is one out of eight possible scenarios for the same background story (e.g., applying to college).

Now given a four-factor causal graph, we can define (4-choose-1 =) 4 inference tasks, where three of the factors are observed and one is inferred. Specifically, we can 
query for the emotion (given the two appraisals and outcome); query for the first or second appraisal (given the other appraisal, outcome, and emotion); or query for the outcome (given both appraisals and the emotion). 

For a given background story (e.g., applying to college), we have 8 scenarios, which could be applied to each of the 4 inference tasks, producing 32 stimuli items. 
For example, we can construct a query for an inference of goal (``Which college did Amy want to go to?''), by specifying the outcome (``Amy got accepted at the local state college, and was rejected from the liberal arts college''), the appraisal of perceived control (``Amy thought that she could influence the decision of the admission process.''), and the emotion (``Amy was joyful.'').

This procedure allows us to flexibly and systematically query for various types of affective inference tasks. 
This pipeline for generating stimuli allows us to scalably and flexibly generate high quality stimuli that are novel. It offers a diverse set of tasks for testing affective reasoning and includes closely matched controls for these tasks, ensuring reliable measurements of capabilities \citep{frank2023baby}.

One natural extension, as mentioned above, is to multimodal reasoning. We include FACS-based generated facial expressions as another possible observed factor in all the tasks, resulting in another 32 (multimodal) questions, which we test in Experiments 2a and 2b (Note that we do not query the models for emotional expression, but simply provide it as another observed factor).

\begin{figure}
    \centering
    \includegraphics[width=0.9\textwidth]{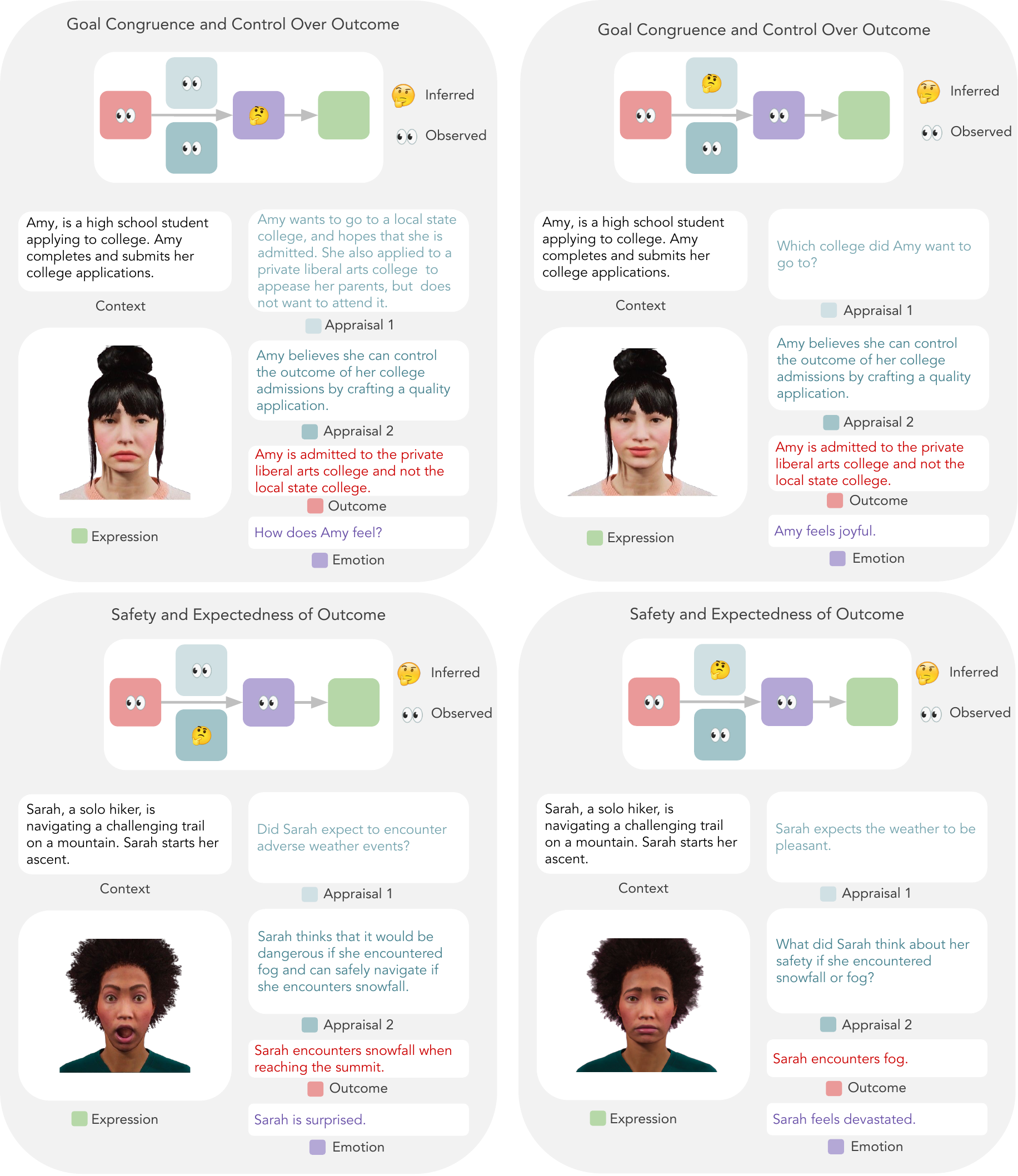}
    \caption{\textbf{Example stimuli used for our experiments.} We can generate stories to ask questions about different affective inferences. 
    Each factor in the causal model, such as appraisals, outcomes, emotions, or expressions, can be varied to elicit different responses. 
    We define different Facial Action Units \citep{ekman1978facial} for different emotions to generate expressions using Unreal Engine. Note that these stimuli are representative of Experiment 2a (top) and 2b (bottom); the corresponding stimuli for Experiments 1a and 1b are text-only, and so would not have the facial expression.
    }
    \label{fig:stimuli}
\end{figure}

Using the same approach, in Experiment 1b and 2b, we generate stimuli based on two additional appraisal dimensions: the \emph{safety} and \emph{expectedness} of the outcome (see \autoref{fig:stimuli} bottom). As with the stimuli for the appraisal dimensions of \emph{goal-congruence} and \emph{perceived-control}, these scenarios are crafted such that the outcome itself is not inherently safe or expected. Instead, these are determined based on the context and the interpretation of the person in the story. 

For Experiments 1a and 2a, we generate 10 background stories for the \emph{goal-congruence} and \emph{perceived-control} appraisal dimensions; and for Experiments 1b and 2b, we generate 10 background stories for the \emph{safety} and \emph{expectedness} dimensions. For each of these stories, we generate 32 text-only stimuli items (Experiments 1a and 1b) and 32 multimodal stimuli items (Experiments 2a and 2b), for a total of 1280 stimuli items.

%% file: sections/02_results.tex
\subsection{Validating our stimuli with human judgments}

We first establish the validity of our stimuli. 
Although our procedural generation pipeline assigns labels to stimuli, based on psychological theory, these may not exactly match human intuition. For instance, people have variability in the intuitive theories they use to make these judgments \citep{saxe2017formalizing, ong2019computational}, and psychological theory may not capture such diversity. 
Thus, we collected responses from 567 native English-speaking participants, averaging about 20 responses per question \footnote{\href{https://osf.io/ajkt6}{https://osf.io/ajkt6}, \href{https://osf.io/7bxwk}{https://osf.io/7bxwk}}, for each of the 1280 questions in our stimuli set. The stimuli are presented in the form of a story, and a question with multiple answer options (\autoref{fig:stimuli}). 
To measure the agreement among participants, we check if an individual's choice matches the majority's choice, calculated without including that individual's response. The agreement score is the average agreement-with-majority-choice across all participants. For the inference task where participants predict emotions, participants choose between four options; random choice would cause the agreement score to be 25\%. On other inference tasks, participants choose between two options; random choice would cause the agreement score to be 50\%.
 
\begin{figure}
    \centering
\includegraphics[width=0.8\textwidth]{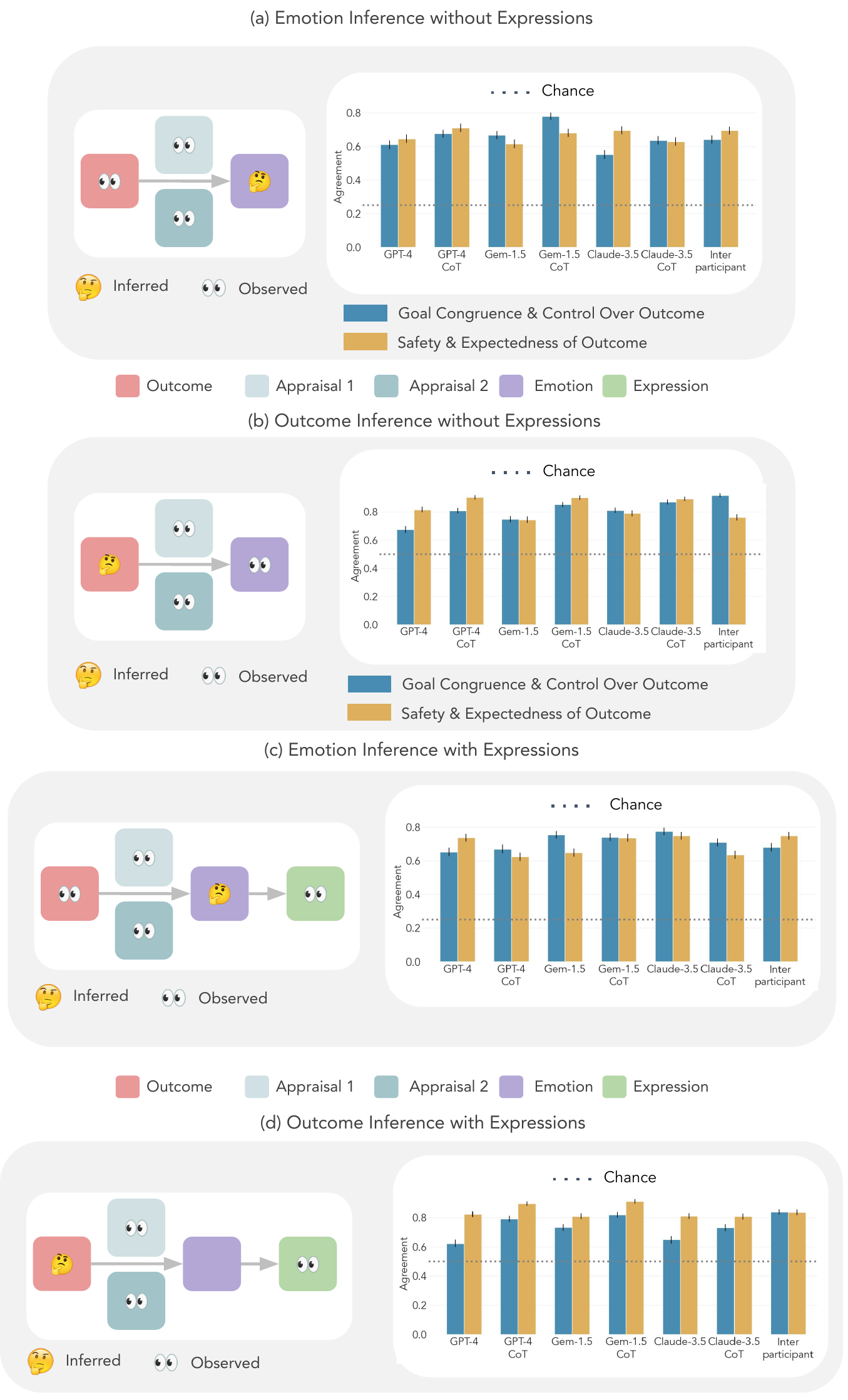}
    \caption{\textbf{Agreement Analysis for Emotion and Outcome Inference.} Inter-participant agreements and model-participant agreements for inferring the (a) emotions and (b) outcomes from the context in Experiments 1a and 1b. (c) and (d): The corresponding agreements for Experiments 2a and 2b, when models and participants were also presented with expressions. Error bars represent 95\% Confidence Intervals.}
    \label{fig:main1}
\end{figure}

We find that the agreement between participants is high, and significantly above what would be expected from random choice (see \autoref{fig:main1}, right-most bars). 
In Experiment 1a (with \emph{goal-conduciveness $\times$ control} scenarios), participants had high agreement when inferring the emotion (63.97\%, 95\% CI = [61.65, 66.29]) given the outcome and appraisals. People also showed high agreement inferring the outcome (91.67\%, 95\% CI = [90.66, 92.68]), the \emph{goal-conduciveness} appraisal (86.09\% [84.82, 87.36]), and the \emph{control} appraisal (72.79\%, 95\% CI = [70.47, 75.11]). These results replicate when instead we considered a different set of appraisals in Experiment 1b (\emph{safety $\times$ expected}): with high rates of agreement for emotion (69.38, 95\% CI= [67.06, 71.70]), outcome (76.10\%, 95\%CI = [74.54, 77.66]), safety (70.31\% [68.63, 71.99]), and expectedness (75.30\%, 95\% CI = [73.06, 77.54]).
%
%
%
%
In Experiment 2a and 2b, stimuli items also had facial expressions, and agreement patterns were also very similar. 
That is, participants can successfully integrate facial expressions with their contexts to make inferences about emotions, outcomes and appraisals. The high agreement shows that our stimuli are able to elicit coherent human judgments, validating their effectiveness.

\begin{figure}
    \centering
    \includegraphics[width=0.8\textwidth]{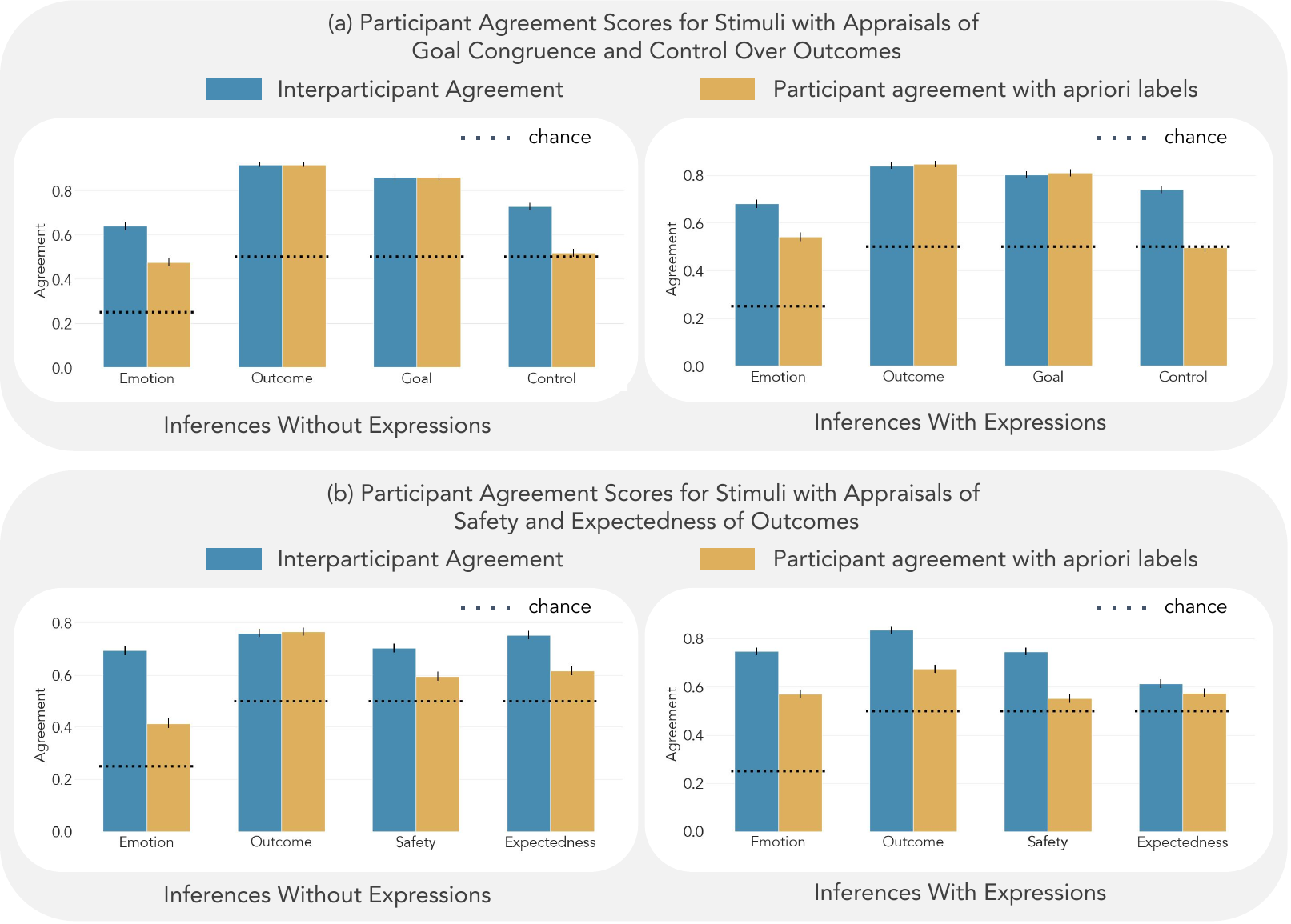}
    \caption{\textbf{Comparison of Inter-participant and Pre-assigned Label Agreement Scores.} Inter-participant agreement scores compared to the agreement scores between participant responses and labels assigned to stimuli prior to collecting human responses. Error bars represent 95\% Confidence Intervals.}
    \label{fig:accres}
\end{figure}

We compared these judgments with the \emph{a priori} labels assigned to the stimuli by our procedural generation pipeline (see \autoref{fig:accres}, \textit{compare blue and yellow}). 
We compute label-participant agreement by comparing the labels to the choices that the majority of participants made for each question. 
Our findings indicate that participants generally predict each other's judgments more accurately than the labels assigned \emph{a priori} during the generation of the stimuli; in other words, the inter-participant agreement is usually the same as or higher than the participant-label agreement. For example, for emotion inference in \emph{safety and expectedness} stimuli items, inter-participant agreement is at 69.38\% [67.67, 71.09], compared to a label-participant agreement of 41.49\% [39.66, 43.32] ($t$=18.05, $p<$0.001). 
This finding underscores the importance of gathering and using participant labels in these subjective affective inference tasks. 
Using psychological theories to assign labels is useful for generating diverse stimuli, but these may diverge substantially from how laypeople judge these stimuli. 
Now, given that we have high inter-participant agreement for our stimuli items, we can now reliably test if model predictions align with human intuitions. 
For the rest of the results presented below, we use the modal participant response (i.e., the majority answer) as the label that we calculate model agreement with, rather than the \emph{a priori} label used by our pipeline.

\subsection{Analyzing affective cognition in foundation models}

To test if contemporary foundation models show human-like patterns in affective reasoning, we prompt three models: \texttt{claude-3.5-sonnet}\footnote{\texttt{claude-3.5-sonnet-20240620}}, \texttt{gpt-4-turbo}\footnote{\texttt{gpt-4-turbo-1106}},  and \texttt{gemini-1.5-pro}\footnote{\texttt{gemini-1.5-pro-002}}. We use two prompting strategies: \texttt{0-shot} and a zero-shot chain-of-thought (CoT), \texttt{0-shot-cot} \citep{kojima2022large, wei2022chain}. We use the most deterministic setting for inference, setting temperature to 0. Additional details and analysis, including specific prompts, other hyperparameter settings and analysis of model and participant response distributions, are available in the supplementary materials. We compare model responses to the choices that the majority of participants made.

In Experiment 1a, when inferring emotions (\autoref{fig:main1} a) from outcomes and appraisals,
we find that model-participant agreement is comparable to interparticipant agreement and significantly above chance. For example, \texttt{claude} has an agreement score of (62.31 \%, 95\% CI=[60.24, 64.38]). 
Agreements scores are generally higher for the \emph{goal-conduciveness $\times$ control} stimuli we used in Experiment 1a, compared to the \emph{safety $\times$ expectedness} stimuli in Experiment 1b. 
%

The foundation models are also excellent at inferring outcomes from emotions and appraisals (79.06\%, 95\% CI = [77.01, 81.12]).
%
For example, for \texttt{gpt-4}, agreement is at 80.88\% [78.83, 82.93] ($t$=14.03, $p<$0.001) for \emph{goal-conduciveness $\times$ control} stimuli and 90.38\% [88.85, 91.91] for \emph{safety $\times$ expectedness} stimuli. 
\texttt{claude}, \texttt{gemini} and \texttt{gpt-4} with CoT again exceed inter-participant agreement scores for the \emph{safety $\times$ expectedness} stimuli; for instance, the agreement scores for \texttt{claude} at 89.20\% [87.59, 90.80] is greater than inter-participant agreement at 76.10\% [73.89, 78.31] ($t$=9.41, $p<$0.001). 
In general, inter-participant agreement for outcomes in Experiment 1a is higher than in Experiment 1b, and so we only observe models being ``better" than humans in Experiment 1b. 

\begin{figure*}[tbh]
    \centering
\includegraphics[width=0.59\textwidth]{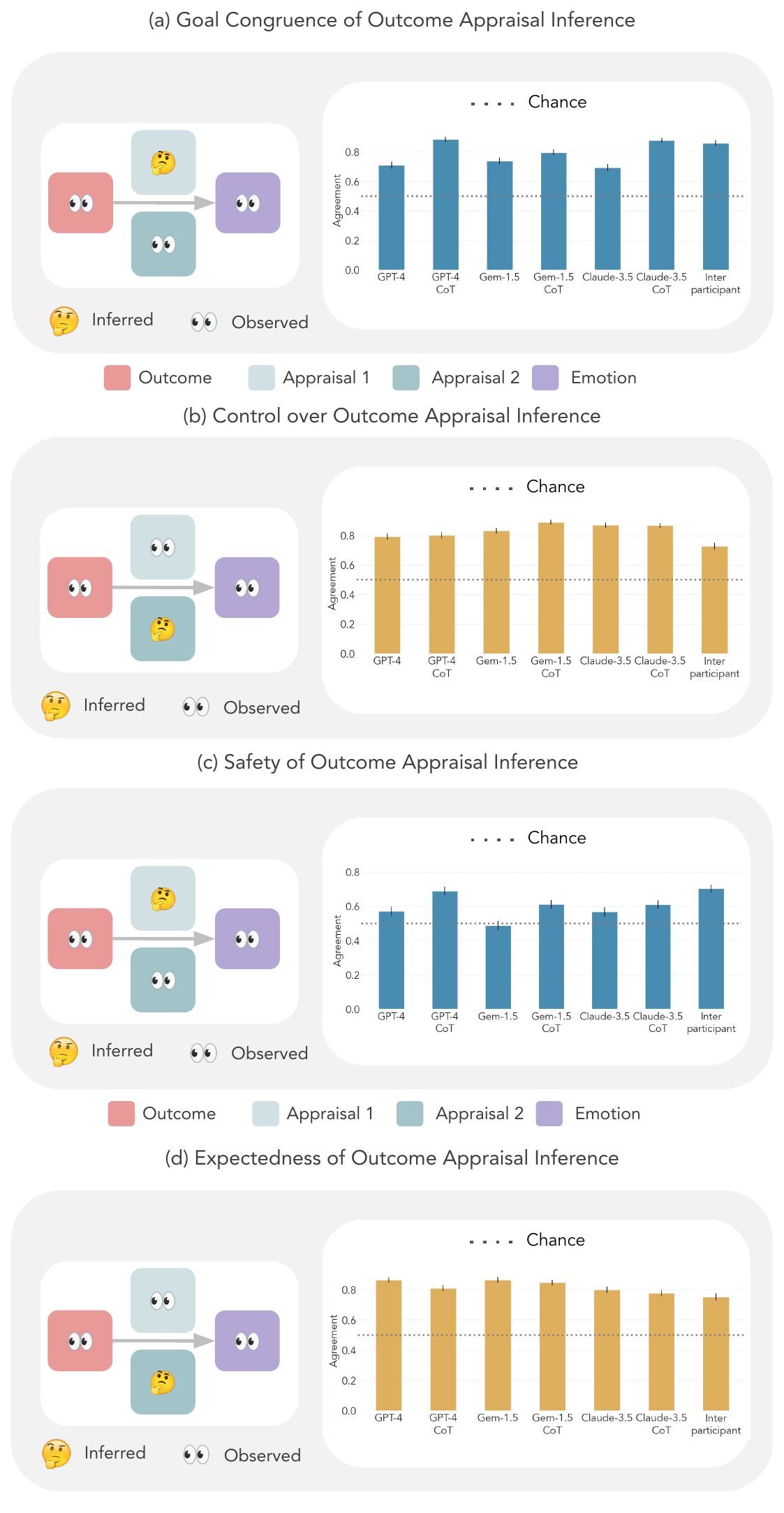}
     \caption{\textbf{Agreement Analysis for Inferring Appraisals from Context without Expressions.} Interparticipant agreements and model participant agreements for inferring the appraisals from the context, for (a, b) Experiment 1a and (c, d) Experiment 1b. Error bars represent 95\% Confidence Intervals.}
    \label{fig:main2}
\end{figure*}

For the task of inferring appraisals from emotions and outcomes (see \autoref{fig:main2}), we see that model-participant agreement scores are highest for predicting the \emph{goal-congruency}, \texttt{gpt-4} has a score of 88.61\% [86.26, 90.96] compared to an interparticipant agreement score of 86.09\% [84.30, 87.88]. Similarly, for \emph{perceived control}, \texttt{claude} has an agreement score of 86.75\% [84.98, 88.52] which is higher than the interparticipant agreement of 72.79\% [70.47, 75.11] ($t$=9.37, $p<$0.001). 
Model agreement scores are much lower for inferring the appraisal of the \emph{safety}: for instance, \texttt{claude} has an agreement score of 60.93\% [58.39, 63.47] compared to the interparticipant agreement of 70.31\% [67.93, 72.69] ($t$=5.28, $p<$ 0.001).

\begin{figure*}[tbh]
    \centering
\includegraphics[width=0.68\textwidth]{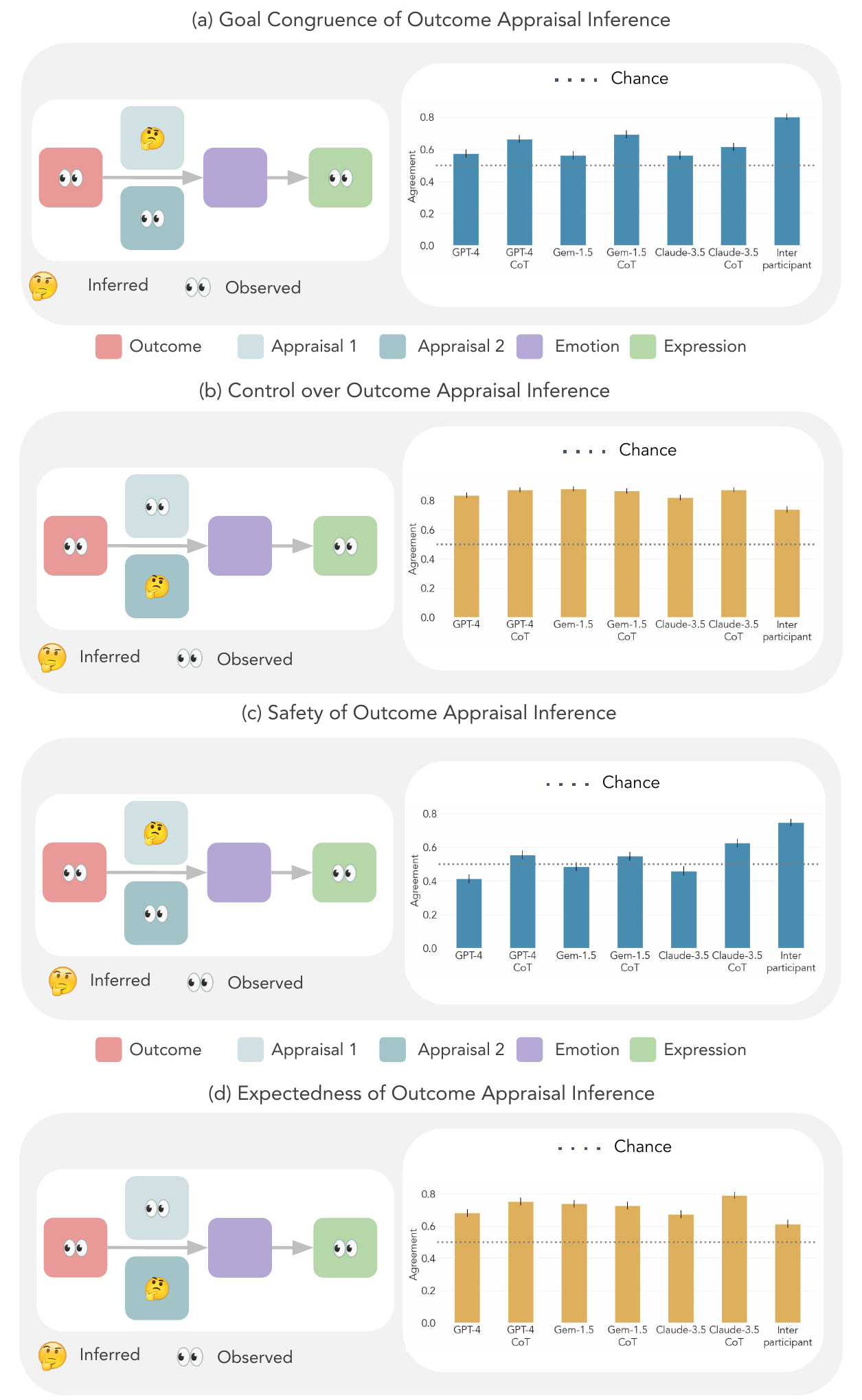}
     \caption{\textbf{Agreement Analysis for Inferring Appraisals from Context with Expressions.} Interparticipant agreements and model participant agreements for inferring the appraisals from the context,  for (a, b) Experiment 2a and (c, d) Experiment 2b. Error bars represent 95\% Confidence Intervals.
     }
    \label{fig:main3}
\end{figure*}

\subsubsection{Incorporating facial expression information.}

When also provided with facial expression stimuli in Experiments 2a and 2b, 
the agreement scores for emotion inference generally increase (a linear model
shows that access to facial expressions increases agreement with human majority
responses by 8.9 percentage points  ($\beta$ = 8.9, 95\% CI: [7.5,10.3]).
For outcome inference (see \autoref{fig:main3}), 
we see that \texttt{claude} is unable to incorporate facial expressions in making inferences about the \emph{goal $\times$ control} stimuli, with the agreement score for outcome inference dropping from 87.01\% [85.26, 88.76] to 73.14\% [70.84, 75.45]  with expressions. 
In contrast, given facial expressions, \texttt{gpt-4} exceeds or matches the scores for inter-participant agreement; for example, for \emph{safety $\times$ expectedness} \texttt{gpt-4} scores 89.57\% [87.98, 91.16] while interparticipant agreement is at 83.58\% [81.65, 85.51], ($t$=4.70, $p<$0.001).
Model participant agreement scores also fall when the task is to infer appraisals from expressions instead of emotions, suggesting that while models are able to integrate expressions when making inferences for outcome and emotions, they struggle while trying to infer appraisals.


\begin{figure*}
    \centering
\includegraphics[width=0.95\textwidth]{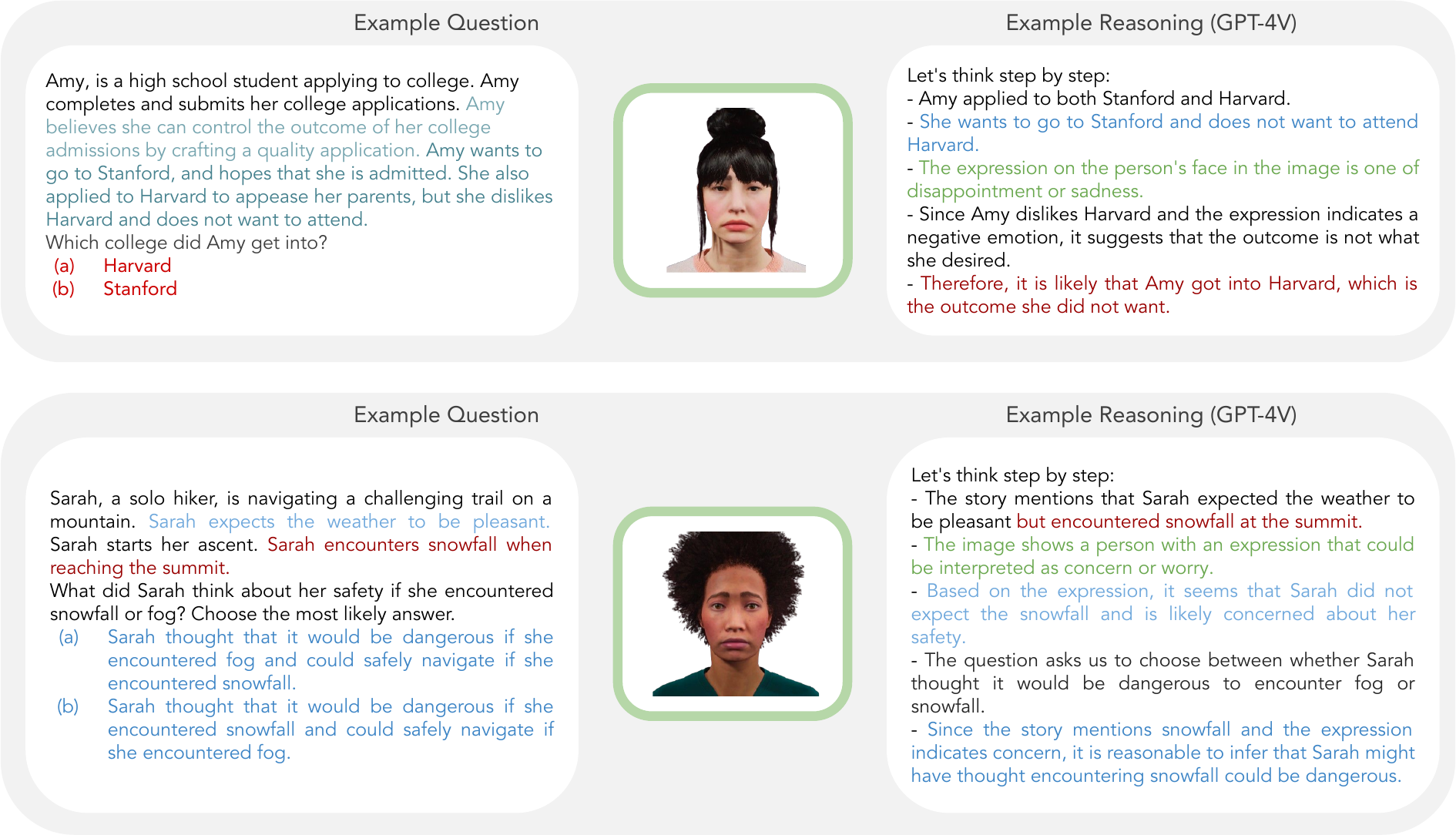}
    \caption{\textbf{Examples of GPT-4 Reasoning Traces Using 0-Shot Chain-of-Thought Prompting.} Examples of a reasoning traces from \texttt{gpt-4} elicited with chain-of-thought prompting, by asking the model to `think step-by-step.'}
    \label{fig:success}
\end{figure*}

\subsubsection{The role of reasoning.}
We find overall that models prompted with chain-of-thought do much better than those without -- a linear model comparing CoT vs. vanilla
prompting across all models shows that CoT increases
agreement with human majority responses by 6.9 percentage points ($\beta$ = 0.069,
95\% CI: [0.062, 0.075]) -- suggesting that reasoning plays a crucial role in improving affective judgment (see \autoref{fig:success} for an example reasoning trace). Across different conditions and models, we find that prompting the model to reason (``thinking step-by-step") before picking an answer increases agreement scores with human modal judgments. For example, for \texttt{gpt-4}, mean agreement for \emph{goal-conduciveness} inferences rises from 71.14\%  [68.79, 73.49] to 88.61\% [86.96, 90.26] ($t$=11.94, $p<$ 0.001), for \texttt{claude}, mean agreement rises from 69.39\% [67.00, 71.78] to 88.05\% [75.19, 79.53] ($t$=11.94, $p<$ 0.001), and for \texttt{gemini} from 74.00\% [71.73, 76.27] to 79.73\% [77.65, 81.81] ($t$=3.64, $p<$ 0.001). Step-by-step reasoning helps the model arrive at inferences that are more aligned with human judgments in many but not all conditions (\autoref{fig:main1}, \autoref{fig:main2}). The improvement in affective judgment through reasoning may mean that as the reasoning capabilities of foundation models improve, so will their affective judgments. 


Finally, to evaluate models' ability to capture human response distributions, we sampled 20 responses per model at temperature 1.0 and compared them to human response distributions using Wasserstein distance, with a uniform distribution serving as a baseline. The analysis (see \autoref{fig:jsd1} and \autoref{fig:jsd2}) revealed that models performed above chance across all tasks, with strongest performance in predicting outcomes, followed by emotions, while showing more variation in appraisal judgments. Models excelled particularly at \emph{goal-conduciveness $\times$ control} judgments and had more difficulty with \emph{safety $\times$ expectedness} judgments. But overall, these results suggesting that foundation models can effectively represent not just modal human responses but full response distributions (see Supplement for more details).

Overall, we find that across our Experiments, foundation models' agreement rates with human modal judgments are high, and match, or sometimes exceed, inter-participant agreements. It does seem that these foundation models are able to integrate information from outcomes, appraisals, emotions, and facial expressions to reason about each of these in turn.

%% file: sections/03_discussion.tex
As AI models continue to advance and become more ``intelligent," we need to define key facets of intelligence, and construct rigorous tests for these facets. Here we build from psychological theory to define  inferences related to affective cognition---reasoning over and understanding people's emotions. Importantly, this goes beyond perception (e.g., detecting sentiment from text, or emotions from facial expressions alone), and involves integrating information from multiple cues (e.g., mental state appraisals, situation outcomes). We propose a framework for systematically constructing a benchmark evaluation dataset for affective cognition in foundation models.  
Our automated evaluation pipeline allowed us to generate diverse and naturalistic stimuli that can be used to systematically, scalably evaluate affective reasoning. 

Using these stimuli, we established human ground-truth, gathering enough judgements for each question to establish the modal human response and agreement of the human population with this response. 
We then performed a comprehensive evaluation of several foundation models (GPT-4, Gemini-1.5, Claude-3), with and without chain-of-thought prompting. 
We found that foundation models tend to agree with human intuitions, matching or in some cases even exceeding inter-participant agreement (predicting the 
modal human judgments better than the average human did. Importantly, we found that chain-of-thought reasoning improved performance, suggesting that improvement in reasoning capabilities of foundation models could lead to even more accurate affective judgments. Finally, we found that some dimensions of appraisal, such as \emph{goal congruency}, were more salient for participants and models compared to others, such as inferences about \emph{perceived control} over the outcome. 
Overall, this suggests that foundation models have acquired the ability to infer emotions in a nuanced way and understand how they influence beliefs and behavior. 

Here we only tested a small number of appraisals, and a small number of background stories (20 stories, which expand to 1280 stimuli items), but our framework is in principle generalizable to a larger number of appraisals (e.g., \citealp{yeo2024meta} identified 47 in the literature) and a potentially infinite number of scenarios. Our framework is also generalizable to other types of social and affective cognition \citep{gandhi2024understanding, fränken2024procedural}. More such research is needed to determine the robustness and limits of these models' mentalizing capabilities.

A number of foundational cognitive questions are raised by these findings:
How is affective cognition represented mechanistically in the weights and activations of the neural network? What types of data are needed for the emergence of emotional reasoning capablities?  
How are these capabilities influenced by post-training alignment versus pretraining on large data? Discovering the origins of affective representations in language models could provide insights and research directions for understanding human cognition \citep{frank2023openly}.

Our work also hints at an interesting future where foundation models might be better at understanding the emotions and mental states of others than the average person. 
We could also envision a future where these models' affective capabilities could be used to develop new approaches for mental health support and intervention \citep{hecht2025using, lee2024large, sharma2023human, zhan2024large}---already, people are finding responses generated by foundation models as more empathic than even those written by trained crisis responders \citep{ovsyannikova2025third, ongai}. 
However, models like these can also introduce potential risks \citep{moore2025expressing}, especially if misused to manipulate or deceive. It is essential to be proactive about measuring capabilities and mitigating the associated risks \citep{ong2021ethical}. Further research around evaluating these models and an ongoing discussion around the ethics of affective computing will be crucial as foundation models continue to improve.

This work presents a rigorous methodology for evaluating affective cognition in both humans and AI systems. We find a broad correspondence between human and model predictions. This points to exciting future directions in using foundation models for interactions requiring emotional understanding.

%% file: sections/14_jsd.tex
To evaluate how well models capture the full distribution of human responses rather than just the most common answer, we generated 20 responses from each model using temperature 1.0 to create response distributions. We then measured the similarity between model and human response distributions using Wasserstein distance (also known as Earth Mover's distance). For comparison, we calculated the Wasserstein distance between human responses and a uniform distribution as a baseline measure of chance performance (``chance'' in \autoref{fig:jsd1} and \autoref{fig:jsd2}). When models refused to answer at temperature 1.0, we assigned equal probability to all possible responses to maintain consistent distribution sizes.

By sampling multiple model responses and comparing their distributions to human answers using Wasserstein distance, we found models performed best at predicting outcomes (\autoref{fig:jsd1}b), followed by emotions (\autoref{fig:jsd1}b), and showed most variation in appraisal judgments (\autoref{fig:jsd2}). While performing above chance across all tasks, models were particularly good at understanding goal congruence and control, but struggled more with safety and expectedness judgments.
These findings suggest that in addition to representing human modal responses, foundation models can represent human response distributions too.


%% file: sections/sup_figs.tex
\begin{figure*}[ht]
\centering
\begin{tcolorbox}[
title={Prompt for 0-shot Evaluation}, width=0.9\textwidth]
\fontsize{7pt}{7pt}\selectfont
\ttfamily
\begin{lstlisting}[language={}]
Answer the questions based on the story. 
Choose your answer from the options provided.
Provide your best guess from the options provided.
Answer precisely in the following format:
A:<option>. <answer>
Example: A:a. cat
\end{lstlisting}
\end{tcolorbox}
\caption{\textbf{Prompt for 0-shot Evaluation.} This prompt is used as the system prompt for 0-shot evaluation of the language model. 
}
\label{exp:0shot}
\end{figure*}

\begin{figure*}[ht]
\centering
\begin{tcolorbox}[
title={Prompt for 0-shot CoT Evaluation}, width=0.9\textwidth]
\fontsize{7pt}{7pt}\selectfont
\ttfamily
\begin{lstlisting}[language={}]
Answer the questions based on the story. 
Choose your answer from the options provided.
Reason step by step before answering in 
`Thought: Let's think step by step:'.
Provide your best guess from the options provided.

Answer in the following format:
Thought: Let's think step by step:
<thought>
A:<option>. <answer>
Example: A:a. cat
\end{lstlisting}
\end{tcolorbox}
\caption{\textbf{Prompt for 0-shot CoT Evaluation.} This prompt is used as the system prompt for 0-shot CoT evaluation of the language model. 
}
\label{exp:0shotcot}
\end{figure*}

\begin{figure*}[ht]
\centering
\begin{tcolorbox}[
title={Prompt for 0-shot Evaluation with Multimodal Stimuli}, width=0.9\textwidth]
\fontsize{7pt}{7pt}\selectfont
\ttfamily
\begin{lstlisting}[language={}]
Answer the questions based on the story and the image.
The image shows the expression that the person in the story feels.
Only pay attention to the expression and 
not the person's physical appearance.
Choose your answer from the options provided.
Provide your best guess from the options provided.
Answer precisely in the following format:
A:<option>. <answer>
Example: A:a. cat
\end{lstlisting}
\end{tcolorbox}
\caption{\textbf{Prompt for 0-shot Evaluation.} This prompt is used as the system prompt for 0-shot evaluation of the model when using stimuli with facial expressions. 
}
\label{exp:0shotx}
\end{figure*}

\begin{figure*}[ht]
\centering
\begin{tcolorbox}[
title={Prompt for 0-shot CoT Evaluation with Multimodal Stimuli}, width=0.9\textwidth]
\fontsize{7pt}{7pt}\selectfont
\ttfamily
\begin{lstlisting}[language={}]
Answer the questions based on the story and the image.
The image shows the expression that the person in the story feels. 
Only pay attention to the expression and 
not the person's physical appearance.
Choose your answer from the options provided.
Reason step by step before answering in 
`Thought: Let's think step by step:'.
Provide your best guess from the options provided.
Answer in the following format:
Thought: 
Let's think step by step:
<thought>
A:<option>. <answer>
Example: A:a. cat
\end{lstlisting}
\end{tcolorbox}
\caption{\textbf{Prompt for 0-shot CoT Evaluation.} This prompt is used as the system prompt for 0-shot CoT evaluation of the model when using stimuli with facial expressions. 
}
\label{exp:0shotcotx}
\end{figure*}

\begin{figure}[tp]
    \centering
\includegraphics[width=\textwidth]{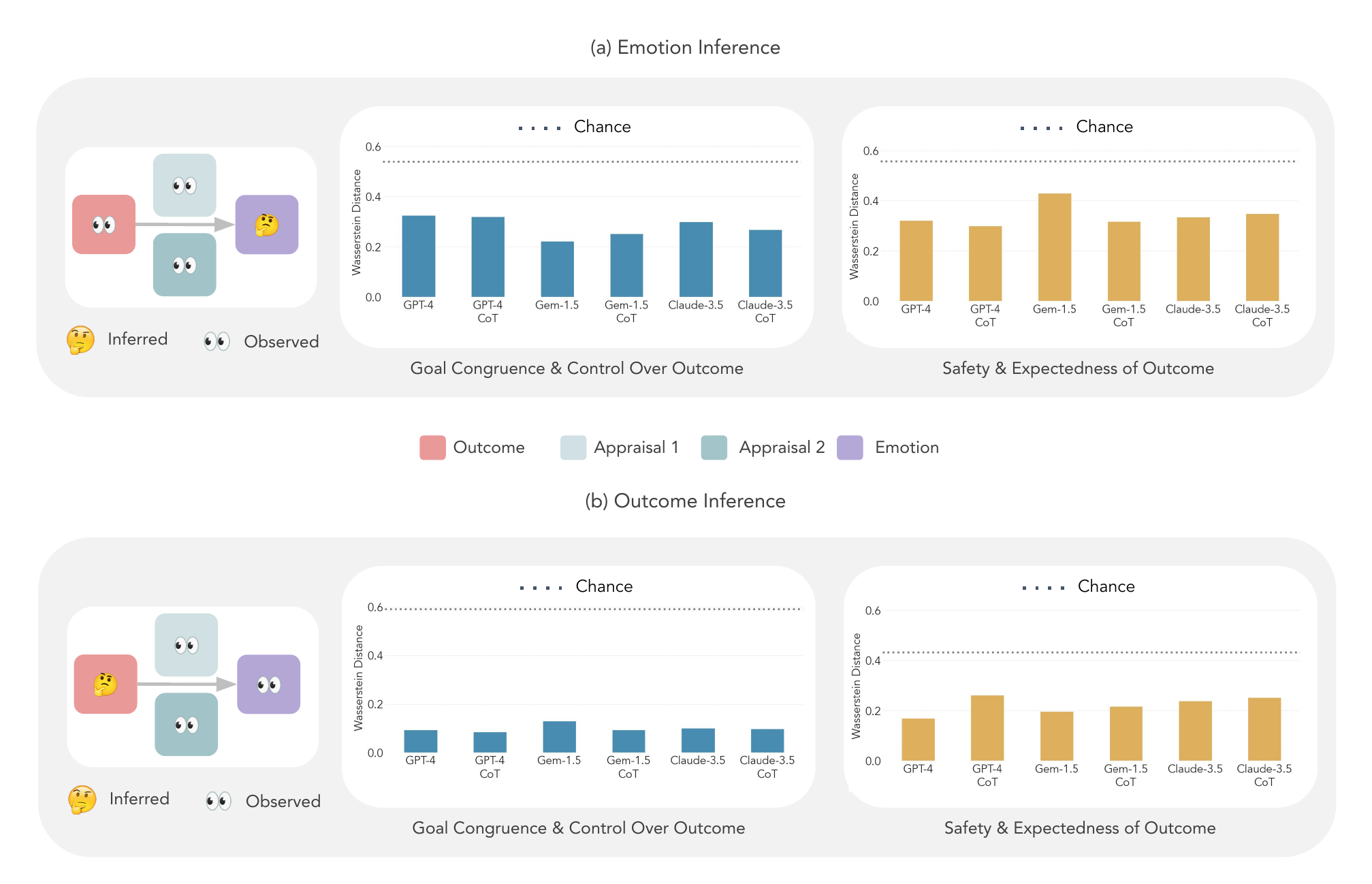}
    \caption{\textbf{Distributional Analysis of Responses for Emotion and Outcome Predictions} Wasserstein Distance (or Earthmover's distance) between model responses and human responses for emotion inference and outcome inference.}
    \label{fig:jsd1}
\end{figure}

\begin{figure}[tp]
    \centering
\includegraphics[width=\textwidth]{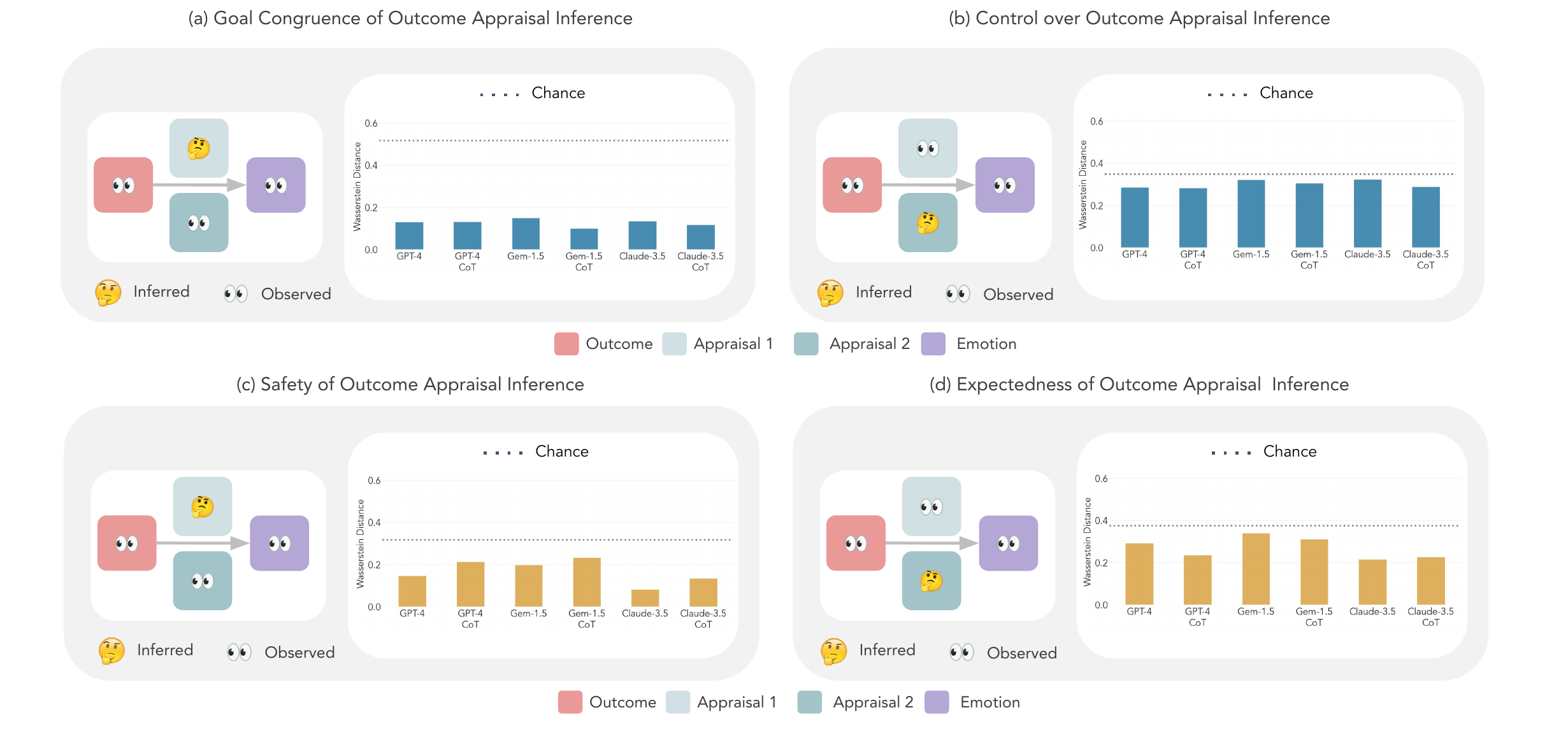}
    \caption{\textbf{Distributional Analysis of Responses for Appraisal Predictions} Wasserstein Distance (or Earthmover's distance) between model responses and human responses for different appraisal predictions.}
    \label{fig:jsd2}
\end{figure}